\documentclass[10pt,twocolumn,letterpaper]{article}

\usepackage{cvpr}
\usepackage{times}
\usepackage{epsfig}
\usepackage{graphicx}
\usepackage{amsmath}
\usepackage{amssymb}

\usepackage{multirow}
\usepackage{array}
\usepackage{subcaption}
\usepackage{bbm}
\usepackage{dsfont}
\usepackage{verbatim}

\usepackage[pagebackref=true,breaklinks=true,colorlinks,bookmarks=false]{hyperref}

\cvprfinalcopy 

\def\cvprPaperID{7062} 

\ifcvprfinal\fi
\begin{document}

\title{Learning Texture Invariant Representation \\ for Domain Adaptation of Semantic Segmentation}

\author{Myeongjin Kim\qquad Hyeran Byun\\Yonsei University\\{\tt\small \{myeongjin.kim,hrbyun\}@yonsei.ac.kr}}
\maketitle
\thispagestyle{empty}

\begin{abstract}
   Since annotating pixel-level labels for semantic segmentation is laborious, leveraging synthetic data is an attractive solution. However, due to the domain gap between synthetic domain and real domain, it is challenging for a model trained with synthetic data to generalize to real data. In this paper, considering the fundamental difference between the two domains as the texture, we propose a method to adapt to the target domain's texture. First, we diversity the texture of synthetic images using a style transfer algorithm. The various textures of generated images prevent a segmentation model from overfitting to one specific (synthetic) texture. Then, we fine-tune the model with self-training to get direct supervision of the target texture. Our results achieve state-of-the-art performance and we analyze the properties of the model trained on the stylized dataset with extensive experiments.
\end{abstract}

\section{Introduction}
Until now, many studies have dealt with semantic segmentation. For supervised semantic segmentation, a large volume of labeled data is required for training. However, the manual annotation for pixel-wise ground truth labels is extremely laborious. For example, it takes 90 min per image to make ground truth label for the Cityscape \cite{cordts2016cityscapes} dataset.

To reduce the cost of annotation, datasets such as GTA5 \cite{richter2016playing} and SYNTHIA \cite{ros2016synthia} are proposed. Since these datasets are generated by computer graphics, the images and pixel-level annotations are automatically generated. However, due to the domain gap between the synthetic domain and the real domain, a model trained with the synthetic data is hard to generalize to the real data. 

Domain adaptation addresses the above issue by reducing the domain gap. One approach is pixel-level adaptation. The pixel-level adaptation uses image translation algorithms like CycleGAN \cite{zhu2017unpaired} to reduce the gap in visual appearance between two domains. Since the synthetic image is translated into the style of the real domain, a model can learn representation for the real domain more easily.

\begin{figure}[t]
    \centering
    \includegraphics[width=0.45\textwidth]{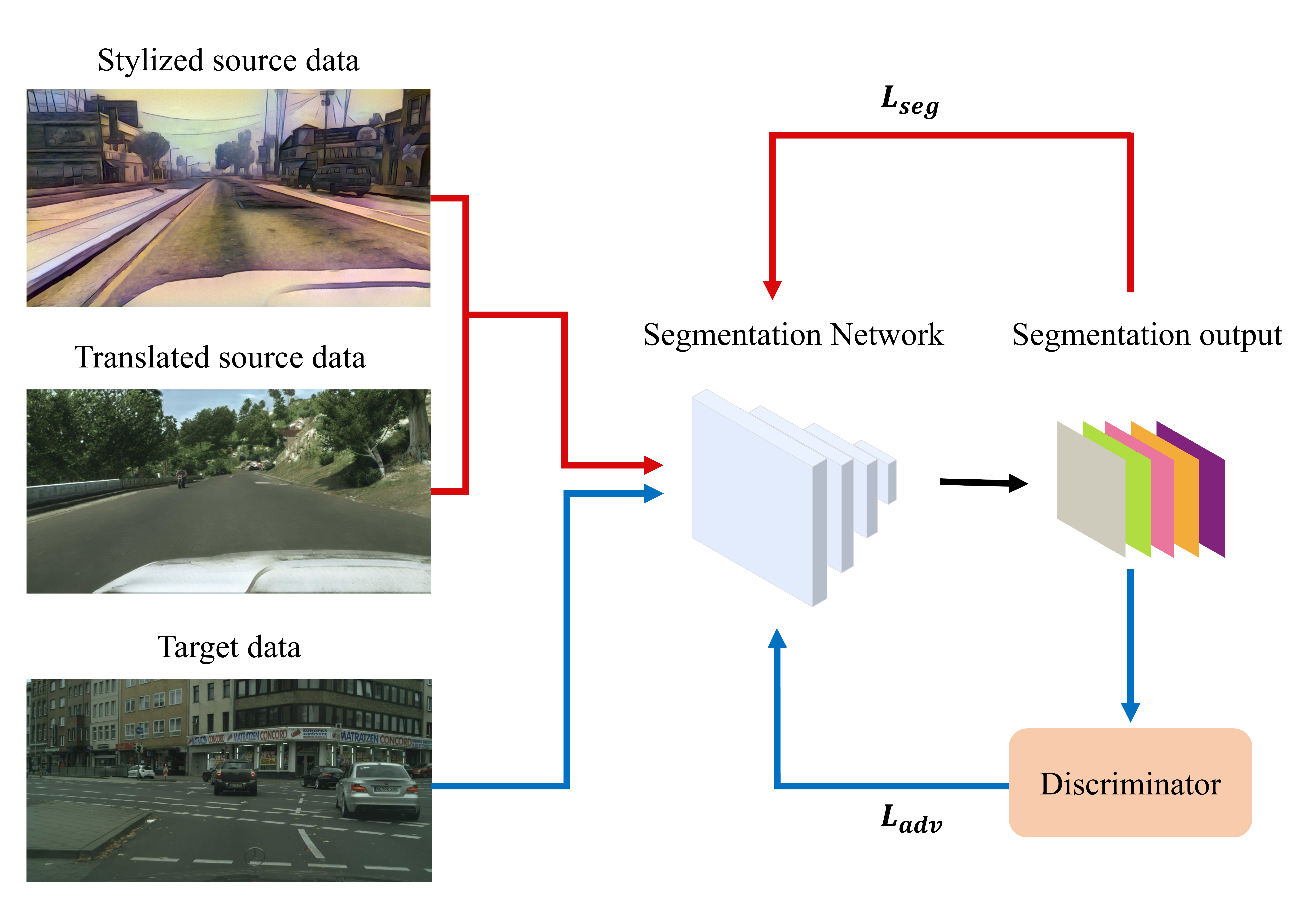}
    \caption{Process of learning texture-invariant representation. We consider both the stylized image and the translated image as the source image. The red line indicates the flow of the source image and the blue line indicates the flow of the target image. By segmentation loss of the stylized source data, the model learns texture-invariant representation. By adversarial loss, the model reduces the distribution gap in feature space.}
    \label{fig:network_figure}
\end{figure}

\begin{figure}[t]
    \centering
    \includegraphics[width=0.3\linewidth]{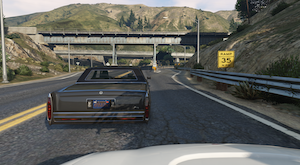}
    \includegraphics[width=0.3\linewidth]{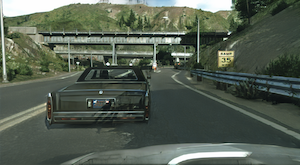}
    \includegraphics[width=0.3\linewidth]{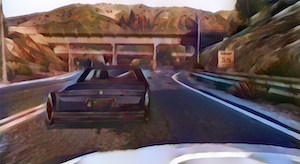}
    \includegraphics[width=0.3\linewidth]{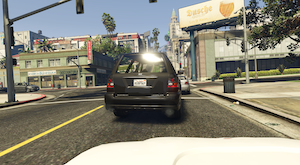}
    \includegraphics[width=0.3\linewidth]{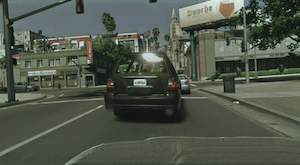}
    \includegraphics[width=0.3\linewidth]{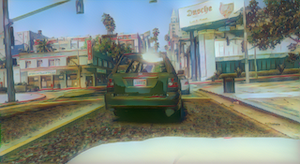}
    \caption{Texture comparison. Original GTA5 \cite{richter2016playing} images (first column), generated images by CycleGAN \cite{zhu2017unpaired} (second column) and by Style-swap \cite{chen2016fast} (third column).}
    \label{fig:texture comparison}
\end{figure}

Although CycleGAN reduces the visual gap between two domains to some extent, overcoming the fundamental difference, the \textit{texture}, is still challenging. In Figure \ref{fig:texture comparison}, the second column shows translated results by CycleGAN. Although the translated images get the Cityscapes' gray color tone, CycleGAN cannot completely translate the synthetic texture into the real texture. Therefore, the possibility of a model to overfit to the synthetic texture still exists. 

To overcome this limitation, we propose a method to adapt to the target domain's texture. First, we generate a texture-diversified source dataset by using a style transfer algorithm. Each source image loses the synthetic texture and gets a random texture. Because of the increased variation of textures, a model trained on the texture-diversified dataset is guided to learn texture-invariant representation. Then, we fine-tune the model using self-training to get direct supervision of the target texture.

Our method achieves state-of-the-art performance on the GTA5 to Cityscapes benchmark. With extensive experiments, we analyze the properties of the model trained on the stylized dataset and compare the differences between ours and CycleGAN-based methods. 

Our contributions are as follows:
\begin{enumerate}
    \item We design a method to adapt to the target domain’s texture for domain adaptation of semantic segmentation, combining pixel-level method and self-training.
    \item We achieve state-of-the-art performance on the GTA5 to Cityscapes benchmark.
    \item With extensive experiments, we analyze the properties of the model trained on the stylized dataset.
    \item We compare our style transfer-based approach and previous CycleGAN-based methods in terms of reducing the domain gap between the synthetic domain and the real domain.
\end{enumerate}


\section{Related Work}
\subsection{Domain adaptation of semantic segmentation}
Domain adaptation transfers knowledge between different domains. Assume two datasets that have similar but different distributions. Let the one which has a larger volume and is more easy to collect as the source domain and the other as the target domain. The goal of domain adaptation is transferring knowledge learned from the source domain to the target domain.

Among some settings of domain adaptation, the unsupervised setting is the most popular, which has access to input data and ground truth labels for the source domain but only input data for the target domain. The goal of unsupervised domain adaptation is to use the fully-labeled source domain properly to improve performance on the unlabeled target domain. Since annotating semantic label is one of the most laborious processes, domain adaptation of semantic segmentation gets much attention recently.

\begin{figure*}[h]
\centering
\begin{tabular}{ccccc}
\begin{subfigure}{0.18\textwidth}
\centering
\includegraphics[height=1.8cm, keepaspectratio]{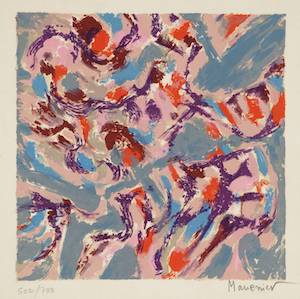}%
\caption{Style}
\end{subfigure}
\hfill
\begin{subfigure}{0.18\textwidth}
\centering
\includegraphics[width=\textwidth]{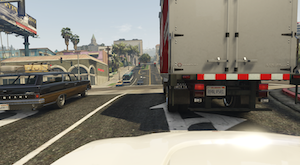}
\caption{Content}
\end{subfigure}
\hfill
\begin{subfigure}{0.18\textwidth}
\centering
\includegraphics[width=\textwidth]{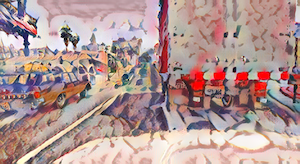}
\caption{AdaIN}
\end{subfigure}
\hfill
\begin{subfigure}{0.18\textwidth}
\centering
\includegraphics[width=\textwidth]{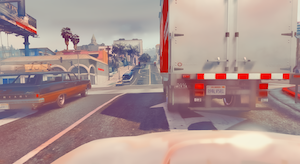}
\caption{PhotoWCT}
\end{subfigure}
\hfill
\begin{subfigure}{0.18\textwidth}
\centering
\includegraphics[width=\textwidth]{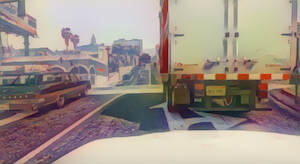}
\caption{Style-Swap}
\end{subfigure}
\end{tabular}
\caption{Results of stylization.}
\label{fig:comparison_stylization}
\end{figure*}

{\bf Pixel-level adaptation.} There exists a visual gap between synthetic and real images, such as texture and lighting. Pixel-level adaptation translates the synthetic source image into the target style using image translation algorithms like CycleGAN \cite{zhu2017unpaired}. Due to the reduced visual gap, a model more easily encodes the representation for the target domain.

{\bf Self-training.} Recently, some works adopt self-training (ST) for domain adaptation of semantic segmentation \cite{zou2018unsupervised,li2019bidirectional}. Generally, ST is applied when labeled training data is scarce. In the unsupervised domain adaptation, because labels of the target domain are absent, it is very attractive to apply ST. \cite{li2019bidirectional} suggests a simple method for self-training. At ST stage, \cite{li2019bidirectional} generates pseudo labels based on the previous model's confident prediction and fine-tune the model with pseudo labels. 

\cite{li2019bidirectional} uses both pixel-level adaptation and self-training. In ablation study, the models trained with ST method outperform other models only using the pixel-level method with a large margin. Considering the fundamental difference between the two domains as the \textit{texture}, powerful performance of ST, which gets direct supervision of the target texture, means that previous methods using pixel-level adaptation are not able to encode the target texture sufficiently.  

Based on this observation, we propose a method that is optimized for encoding the target domain's texture.

\subsection{Style transfer}
Starting from texture synthesis \cite{gatys2015texture} and going through \cite{gatys2016image}, many studies have been conducted about style transfer. Based on the observation that style(texture) and content can be separated, modeling feature statistics makes possible to synthesize image with one image's content and another image's texture. 

Our purpose is, using various textures as a regularizer preventing a model from overfitting to one specific texture, to make the segmentation model learn texture-invariant representation.

\subsection{Texture and shape}
According to recent research \cite{geirhos2018imagenet}, human recognition is based on shape but the ImageNet \cite{deng2009imagenet} pre-trained CNN's criterion is based on texture. To overcome texture-dependency, \cite{geirhos2018imagenet} generates Stylized ImageNet (SIN) using the AdaIN \cite{huang2017arbitrary} style transfer algorithm. Stylized ImageNet lose natural texture and get the various random texture. Since a model trained on SIN cannot predict results based on the local texture, it is enforced to consider the overall structure of the input. \cite{geirhos2018imagenet} demonstrates with experiments that CNN trained on SIN is more shape-dependent like humans and the shape-dependent model is better at classification and detection tasks. 

Inspired by this work, we apply this method to domain adaptation of semantic segmentation task, where the texture is fundamental differences between synthetic and real domains.

\section{Method}
In this section, we present a process for generating texture-diversified datasets and a method to adapt to the target texture. We first diversity the texture of the original source dataset with a style transfer algorithm Style-swap \cite{chen2016fast} and translate the original source dataset with an image translation algorithm CycleGAN \cite{zhu2017unpaired}. Then, our model goes through two training stages.

Stage 1: We train a segmentation model with the texture-diversified dataset to learn texture-invariant representation.

Stage 2: Based on the texture-invariant representation, we fine-tune the model to the target domain's texture.

\begin{figure*}[h]
    \centering
    \includegraphics[width=1\textwidth]{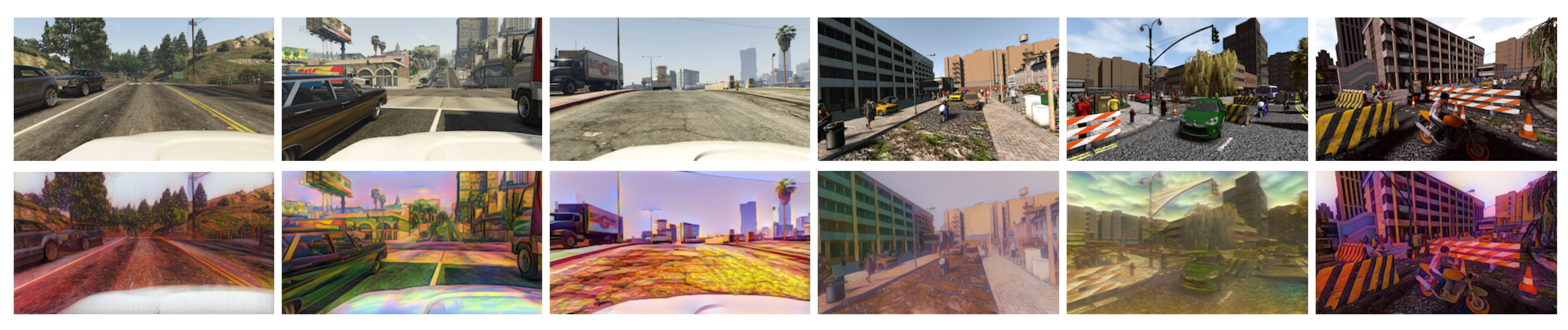}
    \caption{Examples of original images and stylized images.}
    \label{fig:StylizedGTA5}
\end{figure*}

\subsection{Stylized GTA5 / SYNTHIA}
Prior works \cite{hoffman2017cycada,li2019bidirectional} use an image translation method CycleGAN \cite{zhu2017unpaired} to reduce the visual gap between the synthetic and real domains. Although the image translation algorithm makes the source image's overall color tone similar to the real image, it cannot completely translate the synthetic texture into the real one.

To overcome this limitation, we take a more fundamental approach which removes the synthetic texture drastically. Inspired by \cite{geirhos2018imagenet}, we generate Stylized GTA5 and Stylized SYNTHIA. Stylized ImageNet \cite{geirhos2018imagenet} is generated by fast AdaIN \cite{huang2017arbitrary} style transfer algorithm. Although AdaIN is efficient in inference, it distorts the structure of content image considerably with some wave patterns. Unlike the classification task, semantic segmentation task requires accurate pixel-level annotations. Thus, we cannot use AdaIN. The photo-realistic style transfer algorithm \cite{li2018closed} is another option, which preserves the precise structure of the original image using a smoothing step after the stylization step. However, due to the smoothing process which is based on the original content image, final results preserve original synthetic texture. Since our purpose is to remove the synthetic texture using a style transfer algorithm, it is not appropriate to apply the photo-realistic algorithm. Our requirements are three-fold. First, enough stylization effect to remove the synthetic texture, while not distorting the structure of the original image too much. Second, due to the large image resolution and the large volume of the synthetic dataset, the stylization process should be time-efficient. Third, to generate diverse stylized results, it should be able to transfer various styles. Considering above conditions, we choose Style-swap \cite{chen2016fast}. We present stylization results from different methods in Figure \ref{fig:comparison_stylization}.

For a style dataset, we used the \textit{Painter by Numbers} dataset which consists of artistic images. Considering the volume of the GTA5 and SYNTHIA dataset, we use the first split, which contains 11,026 images. The stylized datasets have the same number of images with the original datasets, i.e. one-to-one mapping.

As shown in Figure \ref{fig:StylizedGTA5}, the stylized images drastically lose the synthetic texture and get various random textures. Since each texture is from a different style image, this variety of texture leads a model to encode texture-invariant representation. In other words, the model can learn shape-dependent representation. 

\subsection{Stage 1}
The goal of the first stage is to learn texture-invariant representation using the texture-diversified dataset. We train the segmentation model with both the stylized images by Style-swap \cite{chen2016fast} and the translated images by CycleGAN \cite{zhu2017unpaired}. At each iteration, the stylized or translated inputs are alternately forwarded due to the limitation of memory. While learning texture-invariant representation with the stylized images, the translated images guide the model toward the target style.

Along with the texture regularization, we additionally use the output-level adversarial training \cite{tsai2018learning} to further align feature space between the two different domains. The process of Stage 1 is shown in Figure \ref{fig:network_figure}.

\subsection{Stage 2}
The goal of the second stage is, based on learned texture-invariant representation, to fine-tune the segmentation network to the target domain's texture. For this purpose, we adopt a self-training method. Following the process of \cite{li2019bidirectional}, we generate pseudo labels with the model trained on Stage 1. Among predictions on the target training images, we set predictions with higher confidence than a threshold(0.9) as pseudo-labels. Then we fine-tune the model with the generated pseudo-labels and translated source images. Now the model is directly supervised by the target domain's texture, the model can learn the representation optimized for the target domain. We apply this process iteratively.


\begin{table*}[h]
\caption{Results on GTA5 to Cityscapes.}
\label{tab:gtatocityscapes}
  \centering
  \resizebox{1\textwidth}{!}{
    \begin{tabular}{cccccccccccccccccccccc}
    \hline
    \multicolumn{22}{c}{GTA5 $\rightarrow$  Cityscapes} \\
    \hline
      Base Model & Method & \rotatebox{90}{road} & \rotatebox{90}{side.} & \rotatebox{90}{buil.} &  \rotatebox{90}{wall} & \rotatebox{90}{fence} & \rotatebox{90}{pole} & \rotatebox{90}{t-light} & \rotatebox{90}{t-sign} &\rotatebox{90}{vege.} & \rotatebox{90}{terr.} & \rotatebox{90}{sky} & \rotatebox{90}{pers.} & \rotatebox{90}{rider} &\rotatebox{90}{car} & \rotatebox{90}{truck} & 
      \rotatebox{90}{bus} & \rotatebox{90}{train} & \rotatebox{90}{motor} & \rotatebox{90}{bike} & \rotatebox{90}{mIoU} \\
      \hline
      \multirow{8}{*}{ResNet101} & AdaptSegNet\cite{tsai2018learning} & 86.5 & 36.0 & 79.9 & 23.4 & 23.3 & 23.9 & 35.2 & 14.8 & 83.4 & 33.3 & 75.6 & 58.5 & 27.6 & 73.7 & 32.5 & 35.4 & 3.9 & 30.1 & 28.1 & 42.4\\
      &CLAN\cite{luo2019taking} & 87.0 & 27.1 & 79.6 & 27.3 & 23.3 & 28.3 & 35.5 & 24.2 & 83.6 &27.4 & 74.2 & 58.6 & 28.0 & 76.2 & 33.1 & 36.7 & \textbf{6.7} & 31.9 &31.4 &43.2 \\
      &ADVENT\cite{vu2019advent} & 87.6 & 21.4 & 82.0 & 34.8 & 26.2 & 28.5 & 35.6 & 23.0 & 84.5 & 35.1 & 76.2 & 58.6 & 30.7 & \textbf{84.8} &34.2 & 43.4 & 0.4 & 28.4 & 35.2 & 44.8\\
      &BDL\cite{li2019bidirectional} & 91.0 & 44.7 & 84.2 & \textbf{34.6} & 27.6 & 30.2 & 36.0 & \textbf{36.0} & 85.0 & \textbf{43.6} & 83.0 & 58.6 & \textbf{31.6} & 83.3 & 35.3 & \textbf{49.7} & 3.3 & 28.8 & 35.6 & 48.5\\
      &SIBAN\cite{luo2019significance} & 88.5 & 35.4 & 79.5 & 26.3 & 24.3 & 28.5 & 32.5 & 18.3 & 81.2 & 40.0 & 76.5 & 58.1 & 25.8 & 82.6 & 30.3 & 34.4 & 3.4 & 21.6 & 21.5 & 42.6\\
      &AdaptPatch\cite{tsai2019domain} & 92.3 & 51.9 & 82.1 & 29.2 & 25.1 & 24.5 & 33.8 & 33.0 & 82.4 & 32.8 & 82.2 & 58.6 & 27.2 & 84.3 & 33.4 & 46.3 & 2.2 & 29.5 & 32.3 & 46.5\\
      &MaxSquare\cite{chen2019domain} & 89.4 & 43.0 & 82.1 & 30.5 & 21.3 & 30.3 & 34.7 & 24.0 & \textbf{85.3} & 39.4 & 78.2 & \textbf{63.0} & 22.9 & 84.6 & \textbf{36.4} & 43.0 & 5.5 & 34.7 & 33.5 & 46.4\\
      &Ours & \textbf{92.9} & \textbf{55.0} & \textbf{85.3} & 34.2 & \textbf{31.1} & \textbf{34.9} & \textbf{40.7} & 34.0 & 85.2 & 40.1 & \textbf{87.1} & 61.0 & 31.1 & 82.5 & 32.3 & 42.9 & 0.3 & \textbf{36.4} & \textbf{46.1} & \textbf{50.2}\\
      \hline
      \multirow{8}{*}{VGG16} & AdaptSegNet\cite{tsai2018learning} & 87.3 & 29.8 & 78.6 & 21.1 & 18.2 & 22.5 & 21.5 & 11.0 & 79.7 & 29.6 & 71.3 & 46.8 & 6.5 & 80.1 & 23.0 & 26.9 & 0.0 & 10.6 & 0.3 & 35.0\\
      &CLAN\cite{luo2019taking} & 88.0 & 30.6 & 79.2 & 23.4 & 20.5 & 26.1 & 23.0 & 14.8 & 81.6 & 34.5 & 72.0 & 45.8 & 7.9 & 80.5 & \textbf{26.6} & 29.9 & 0.0 & 10.7 & 0.0 & 36.6 \\
      &ADVENT\cite{vu2019advent} & 86.8 & 28.5 & 78.1 & 27.6 & 24.2 & 20.7 & 19.3 & 8.9 & 78.8 & 29.3 & 69.0 & 47.9 & 5.9 & 79.8 & 25.9 & \textbf{34.1} & 0.0 & 11.3 & 0.3 & 35.6\\
      &BDL\cite{li2019bidirectional} & 89.2 & 40.9 & 81.2 & 29.1 & 19.2 & 14.2 & 29.0 & \textbf{19.6} & 83.7 & 35.9 & 80.7 & \textbf{54.7} & \textbf{23.3} & \textbf{82.7} & 25.8 & 28.0 & 2.3 & \textbf{25.7} & \textbf{19.9} & 41.3\\
      &SIBAN\cite{luo2019significance} & 83.4 & 13.0 & 77.8 & 20.4 & 17.5 & 24.6 & 22.8 & 9.6 & 81.3 & 29.6 & 77.3 & 42.7 & 10.9 & 76.0 & 22.8 & 17.9 & 5.7 & 14.2 & 2.0 & 34.2 \\
      &AdaptPatch\cite{tsai2019domain} & 87.3 & 35.7 & 79.5 & 32.0 & 14.5 & 21.5 & 24.8 & 13.7 & 80.4 & 32.0 & 70.5 & 50.5 & 16.9 & 81.0 & 20.8 & 28.1 & 4.1 & 15.5 & 4.1 & 37.5\\
      &DRPC\cite{yue2019domain} & 84.6 & 31.5 & 76.3 & 25.4 & 17.2 & 28.2 & 21.5 & 13.7 & 80.7 & 26.8 & 74.9 & 47.5 & 15.8 & 77.1 & 22.2 & 22.7 & 1.7 & 8.9 & 9.7 & 36.1\\
      &Ours & \textbf{92.5} & \textbf{54.5} & \textbf{83.9} & \textbf{34.5} & \textbf{25.5} & \textbf{31.0} & \textbf{30.4} & 18.0 & \textbf{84.1} & \textbf{39.6} & \textbf{83.9} & 53.6 & 19.3 & 81.7 & 21.1 & 13.6 & \textbf{17.7} & 12.3 & 6.5 & \textbf{42.3}\\
      \hline
    \end{tabular}
 }
\end{table*}


\begin{table*}[h]
\caption{Results on SYNTHIA to Cityscapes.}
\label{tab:synthia2cityscapes}
  \centering
  \resizebox{1\textwidth}{!}{
    \begin{tabular}{cccccccccccccccc}
    \hline
    \multicolumn{16}{c}{SYNTHIA $\rightarrow$ Cityscapes} \\
    \hline
      Base Model & Method & \rotatebox{90}{road} & \rotatebox{90}{side.} & \rotatebox{90}{buil.} & \rotatebox{90}{t-light} & \rotatebox{90}{t-sign} &\rotatebox{90}{vege.} & \rotatebox{90}{sky} & \rotatebox{90}{pers.} & \rotatebox{90}{rider} &\rotatebox{90}{car} & 
      \rotatebox{90}{bus} & \rotatebox{90}{motor} & \rotatebox{90}{bike} & \rotatebox{90}{mIoU} \\
      \hline
      \multirow{8}{*}{ResNet101} & AdaptSegNet\cite{tsai2018learning} & 84.3 & 42.7 & 77.5 &  4.7 & 7.0 & 77.9 & 82.5 & 54.3 & 21.0 & 72.3 & 32.2 & 18.9 & 32.3 & 46.7\\
      &CLAN\cite{luo2019taking} & 81.3 & 37.0 & 80.1 & 16.1 & 13.7 & 78.2 & 81.5 & 53.4 & 21.2 & 73.0 & 32.9 & 22.6 & 30.7 & 47.8\\
      &ADVENT\cite{vu2019advent} & 85.6 & 42.2 & 79.7 & 5.4 & 8.1 & 80.4 & 84.1 & 57.9 &23.8 & 73.3 & 36.4 & 14.2 & 33.0 & 48.0\\
      &BDL\cite{li2019bidirectional} & 86.0 & 46.7 & \textbf{80.3} & 14.1 & 11.6 & 79.2 & 81.3 & 54.1 & \textbf{27.9} & 73.7 & \textbf{42.2} & \textbf{25.7} & \textbf{45.3} & \textbf{51.4}\\
      &SIBAN\cite{luo2019significance} & 82.5 & 24.0 & 79.4 & \textbf{16.5} & 12.7 & 79.2 & 82.8 & \textbf{58.3} & 18.0 & 79.3 & 25.3 & 17.6 & 25.9 & 46.3\\
      &AdaptPatch\cite{tsai2019domain} & 82.4 & 38.0 & 78.6 & 3.9 & 11.1 & 75.5 & \textbf{84.6} & 53.5 & 21.6 & 71.4 & 32.6 & 19.3 & 31.7 & 46.5\\
      &MaxSquare\cite{chen2019domain} & 82.9 & 40.7 & 80. & 12.8 & \textbf{18.2} & \textbf{82.5} & 82.2 & 53.1 & 18.0 & 79.0 & 31.4 & 10.4 & 35.6 & 48.2\\
      &Ours & \textbf{92.6} & \textbf{53.2} & 79.2 & 1.6 & 7.5 & 78.6 & 84.4 & 52.6 & 20.0 & \textbf{82.1} & 34.8 & 14.6 & 39.4 & 49.3\\
      \hline
      \multirow{7}{*}{VGG16} & AdaptSegNet\cite{tsai2018learning} & 78.9 & 29.2 & 75.5 & 0.1 & 4.8 & 72.6 & 76.7 & 43.4 & 8.8 & 71.1 & 16.0 & 3.6 & 8.4 & 37.6\\
      &CLAN\cite{luo2019taking} & 80.4 & 30.7 & 74.7 & 1.4 & 8.0 & 77.1 & 79.0 & \textbf{46.5} & 8.9 & 73.8 & 18.2 & 2.2 & 9.9 & 39.3\\
      &ADVENT\cite{vu2019advent} & 67.9 & 29.4 & 71.9 & 0.6 & 2.6 & 74.9 & 74.9 & 35.4 & 9.6 & 67.8 & 21.4 & 4.1 & 15.5 & 36.6\\
      &SIBAN\cite{luo2019significance} & 70.1 & 25.7 & \textbf{80.9} & 3.8 & 7.2 & 72.3 & 80.5 & 43.3 & 5.0 & 73.3 & 16.0 & 1.7 & 3.6 & 37.2 \\
      &AdaptPatch\cite{tsai2019domain} & 72.6 & 29.5 & 77.2 & 1.4 & 7.9 & 73.3 & 79.0 & 45.7 & 14.5 & 69.4 & 19.6 & 7.4 & 16.5 & 39.6\\
      &DRPC\cite{yue2019domain} & 77.5 & 30.7 & 78.6 & \textbf{10.6} & \textbf{16.1} & 75.2 & 76.5 & 44.1 & \textbf{15.8} & 69.9 & 14.7 & \textbf{8.6} & 17.6 & 41.2\\
      &Ours & \textbf{89.8} & \textbf{48.6} & 78.9 & 0.0 & 4.7 & \textbf{80.6} & \textbf{81.7} & 36.2 & 13.0 & \textbf{74.4} & \textbf{22.5} & 6.5 & \textbf{32.8} & \textbf{43.8}\\
      \hline
    \end{tabular}
    }
\end{table*}

\subsection{Training objective}
{\bf Segmentation model training.}
Since the ground truth label is only available in the source domain, the segmentation loss is defined as:

\begin{equation}
L_{seg}(I_s) = -\sum_{h,w}\sum_{c=1}^Cy_s^{h,w,c}\log P_s^{(h,w,c)}
\end{equation}

And when the target image is given, we calculate the adversarial loss using discriminator.

\begin{equation}
L_{adv}(I_t) = -\sum_{h,w}\log D(P_t^{(h,w,c)})
\end{equation}

where $I_s$ and $I_t$ are the input images from the source domain and the target domain. $P_s^{(h,w,c)}$ and $P_t^{(h,w,c)}$ are the final feature of the source and target image. $y_s^{h,w,c}$ is the source domain's ground truth pixel label. $C$ is the number of classes and $D$ is a fully convolutional discriminator.

Therefore, the total loss function for the segmentation network is defined as:
\begin{equation}
L(I_s,I_t) = L_{seg}(I_s) + \lambda_{adv} L_{adv}(I_t)
\end{equation}

{\bf Discriminator Training.} 
The discriminator takes source and target features and classifies whether it is from the source or target domain. 

\begin{equation}
\begin{split}
L_D(P) & = -\sum_{h,w}((1-z)\log D(P_s^{(h,w,c)})\\
& + z\log D(P_t^{(h,w,c)}))
\end{split}
\end{equation}
where $z = 0$ if the feature is from source domain and $z = 1$ if the feature is from target domain.

{\bf Self-training.} 
In stage 2, to get direct supervision of the target domain's texture, we calculate the segmentation loss for generated pseudo-labels in target images.

\begin{equation}
L_{ST}(I_t) = -\sum_{h,w}\mathbbm{1}_{pseudo}\sum_{c=1}^C\hat{y}_t^{h,w,c}\log P_t^{(h,w,c)}
\end{equation}
where $\mathbbm{1}_{pseudo}$ indicates whether each pixel of the target training set is pseudo-label or not.


\section{Experiments}
{\bf Dataset.} GTA5 \cite{richter2016playing} is a dataset which contains 24,966 synthetic images from the video game with $1914 \times 1052$ resolution. The semantic labels are compatible with the Cityscapes dataset in 19 classes.

For SYNTHIA \cite{ros2016synthia}, we use the SYNTHIA-RAND-CITYSCAPES partition with 9,400 images of $1280 \times 760$ resolution. We validate on 13 common classes with the Cityscapes dataset.

Cityscapes \cite{cordts2016cityscapes} is a dataset which contains 5,000 densely annotated images with $2048 \times 1024$ resolution. We use 2,975 training images and 500 validation images.

{\bf Network architecture.} We use the DeepLab-v2 \cite{chen2017deeplab} model with ResNet-101 \cite{he2016deep} and VGG-16 \cite{simonyan2014very} which are pretrained on ImageNet \cite{deng2009imagenet}. For the discriminator, we adopt similar architecture to \cite{radford2015unsupervised}. The network contains 5 convolution layers with $4 \times 4$ kernel size, channel numbers are \{64,128, 256, 512, 1\} and stride of 2.

{\bf Training detail.} We implement our experiment using the Pytorch library on a single GTX 1080 Ti. To optimize the segmentation model, we use the SGD method. The momentum is set as 0.9. The initial learning rate is $1.0 \times 10^{-4}$ for Stage 1. Due to the variation of the stylized dataset, a high learning rate makes training unstable. Therefore, we set smaller value than prior works which adopt the same architecture \cite{tsai2018learning,luo2019taking,vu2019advent,chang2019all,li2019bidirectional}. The same learning rate is used for fine-tuning in Stage 2. For the learning rate schedule, we adopt the polynomial procedure mentioned in \cite{chen2017deeplab}. For optimizing discriminator, we use Adam for optimizing method with the learning rate $1.0 \times 10^{-4}$ and the momentum 0.9 and 0.99. We set $\lambda_{adv}$ as 0.001. Inputs are resized to $1024 \times 512$.

{\bf Comparison with state-of-the-art models.} 
As shown in Table \ref{tab:gtatocityscapes}, our method outperforms all previous state-of-the-art methods on GTA5-to-Cityscapes. BDL \cite{li2019bidirectional} iterates the training process six times and outperforms other models with a large margin. Our model surpasses the performance of BDL with only two iterations of the segmentation training as shown in Table \ref{tab:effect of self-training}. These results show that our method (first learn texture-invariant representation, then fine-tune toward target texture) is more effective than a simple self-training method.

For the SYNTHIA to Cityscapes, we compare methods that evaluate performance on 13 classes in Table \ref{tab:synthia2cityscapes}. Our method shows outstanding performance in classes like \textit{road} and \textit{sidewalk}, which occupy large area in input images. Since large-area classes will be more affected by texture, our texture-based method outperforms others in these classes. 

Results also report our performance on small classes like \textit{t-light}, \textit{t-sign} and \textit{person} are lower than other methods. Although the texture is a fundamental difference between the synthetic and real domains, it is not the only factor causing the domain gap. The layout gap is also an important factor that we didn’t handle in this paper. This layout gap brings discrepancy of shape distribution across domains. In SYNTHIA, \textit{t-light}, \textit{t-sign}, and \textit{person} are depicted much smaller compared to GTA5 and Cityscapes. Since the shape is more decisive factors than texture for small-area classes, our shape-dependent representation, which is fitted to SYNTHIA's shape distribution, is hard to be transferred to Cityscapes' shape distribution. 

Also as quantitatively shown in \cite{wrenninge2018synscapes}, the domain gap between SYNTHIA and Cityscapes is much larger than the domain gap between GTA5 and Cityscapes, especially for \textit{t-light} and \textit{t-sign}. Other methods use an additional technique like class-ratio prior \cite{vu2019advent} to reduce the layout gap.

{\bf Comparison of class-wise performance.} We provide the basis for the above claim through a class-wise ablation study. In Table \ref{tab:large_small}, IoUs are from large (texture-sensitive) and small (texture-insensitive) classes in the Stage 1. Models trained on \textit{Stylized} dataset outperform models trained on \textit{Translated} and \textit{Original} dataset in large-area classes like \textit{road} and \textit{sidewalk}. Among other large-area classes, since \textit{road} and \textit{sidewalk} have similar layout distribution, texture is an especially important factor for these classes.

On the other hand, \textit{Original} outperforms other methods in \textit{t-light} and \textit{t-sign}. \cite{wrenninge2018synscapes} shows, when using the synthetic and real data together, performance increases significantly in \textit{t-light} and \textit{t-sign} compared to other classes. This means texture is not a decisive factor for these classes and the sharp original image is more helpful for improving performance in the real domain.

\begin{table}[h!]
\caption{Ablation study on large \& small classes.}
\label{tab:large_small}
  \centering
  \resizebox{0.45\textwidth}{!}{
    \begin{tabular}{cccccc}
    \hline
    \multicolumn{6}{c}{SYNTHIA $\rightarrow$ Cityscapes} \\
    \hline
       Base Model & Source Type & \rotatebox{90}{road} & \rotatebox{90}{side.} & \rotatebox{90}{t-light} & \rotatebox{90}{t-sign} \\
      \hline
      \multirow{3}{*}{ResNet101} 
      &Stylized & \textbf{87.7} & \textbf{44.1} & 1.0 & 5.8\\ 
      &Translated & 84.6 & 40.6 & 1.3 & 5.0\\
      &Original \cite{tsai2018learning} & 79.2& 37.2 & \textbf{9.9} & \textbf{10.5}\\
      \hline
      \multirow{3}{*}{VGG16} 
      &Stylized & \textbf{86.1} & \textbf{36.4} & \textbf{0.3} & 1.7 \\
      &Translated & 75.6 & 31.9 & 0 & 3.6 \\
      &Original \cite{tsai2018learning} & 78.9& 29.2 & 0.1 & \textbf{4.8} \\
      \hline
    \end{tabular}
  }
\end{table}

\section{Discussion}
\subsection{Comparison with CycleGAN-based methods}
In this section, we compare the differences between ours and CycleGAN-based methods.

First, CyCADA \cite{hoffman2017cycada} uses CycleGAN to reduce the visual gap between the synthetic and real domains. However, while CycleGAN's generator is trained to generate undistinguishable images from the target domain, CycleGAN is prone to generate inappropriate images. 

\begin{figure}[h!]
    \centering
    \includegraphics[width=0.5\textwidth]{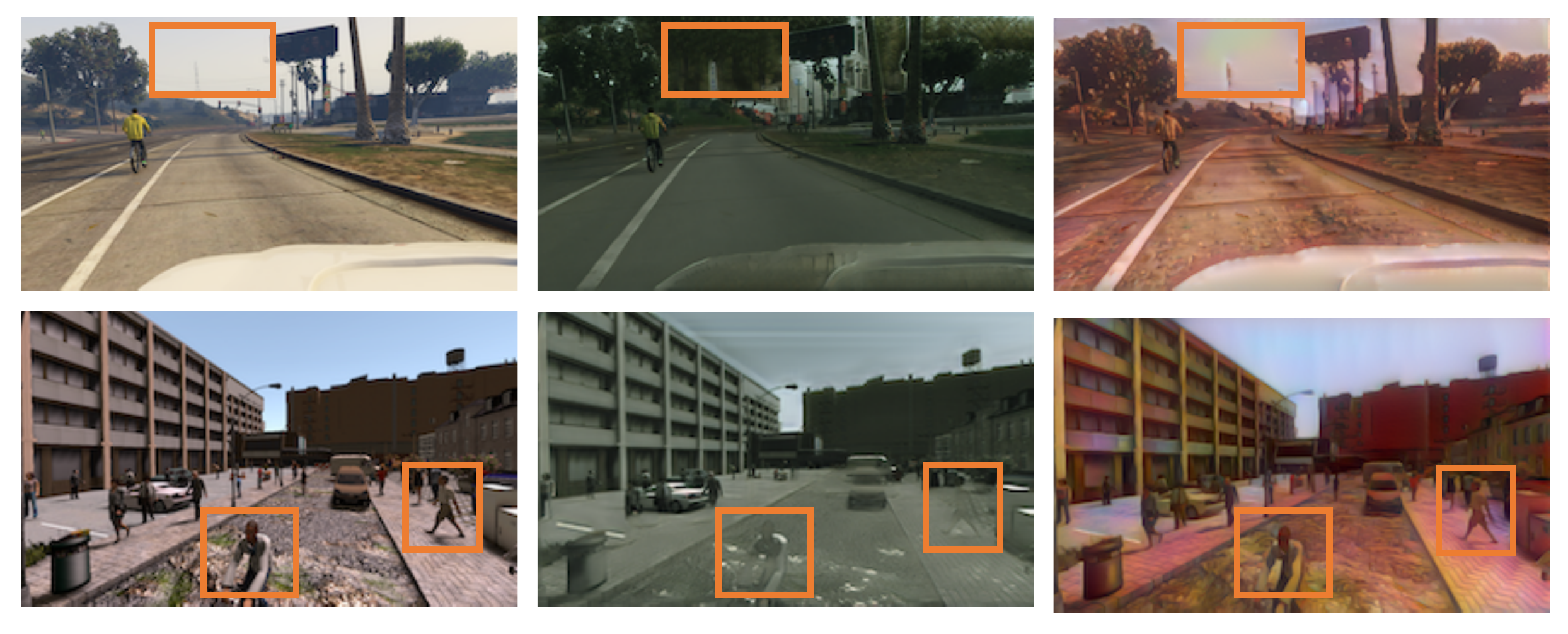}
    \caption{Inappropriate generation of CycleGAN. Original images (first column), generated images by CycleGAN (second column) and Style-swap (third column).}
    \label{fig:falsely_generated_images}
\end{figure}

In Figure \ref{fig:falsely_generated_images}, for GTA5 to Cityscapes (first row), CycleGAN generates \textit{vegetation}-like artifact on the \textit{sky} to match Cityscapes' distribution. For SYNTHIA to Cityscapes (second row), CycleGAN blurs out \textit{person} to match Cityscapes' color distribution. 
Despite CycleGAN discriminator's PatchGAN structure, these patterns are easily observed. On the other hand, because Style-swap transfers style based on local patch, Style-swap doesn't show such patterns.

\begin{figure*}[t!]
    \centering
    \includegraphics[width=1\textwidth]{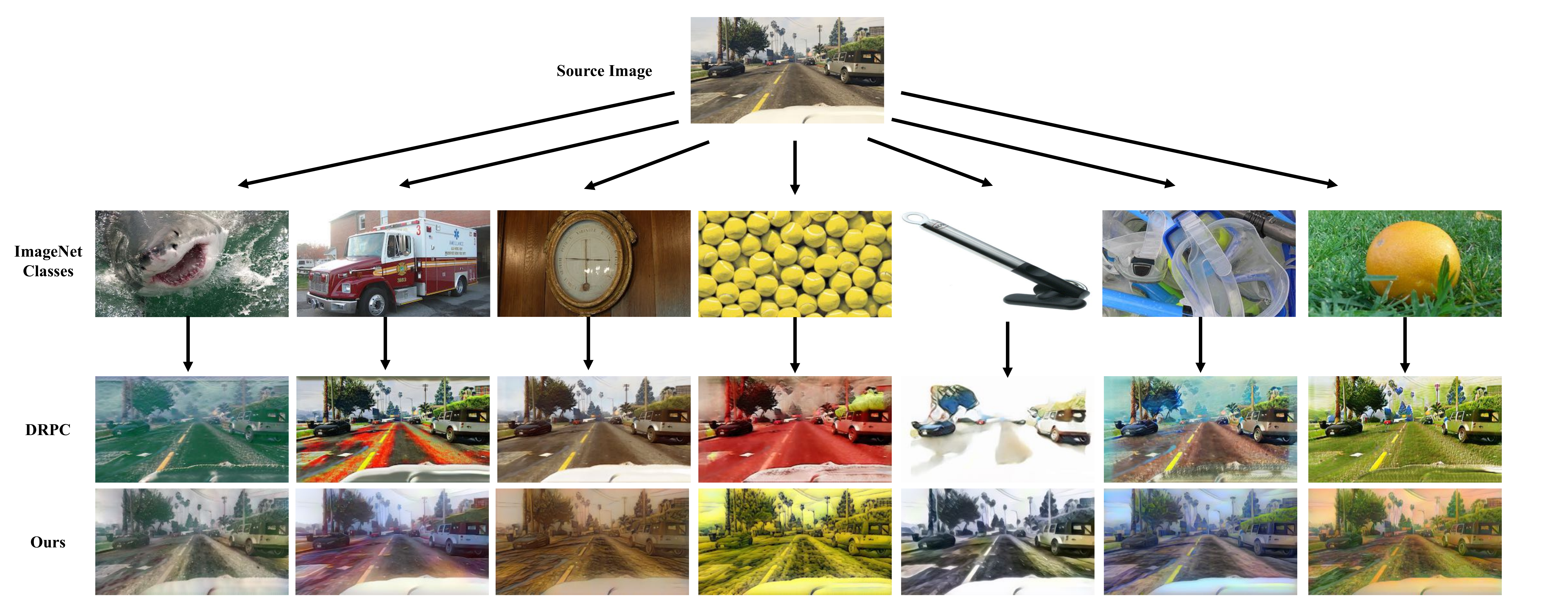}
    \caption{Stylization comparison with DRPC.}
    \label{fig:DRPC_images}
\end{figure*}

\begin{figure*}[h!]
    \centering
    \includegraphics[width=0.95\textwidth]{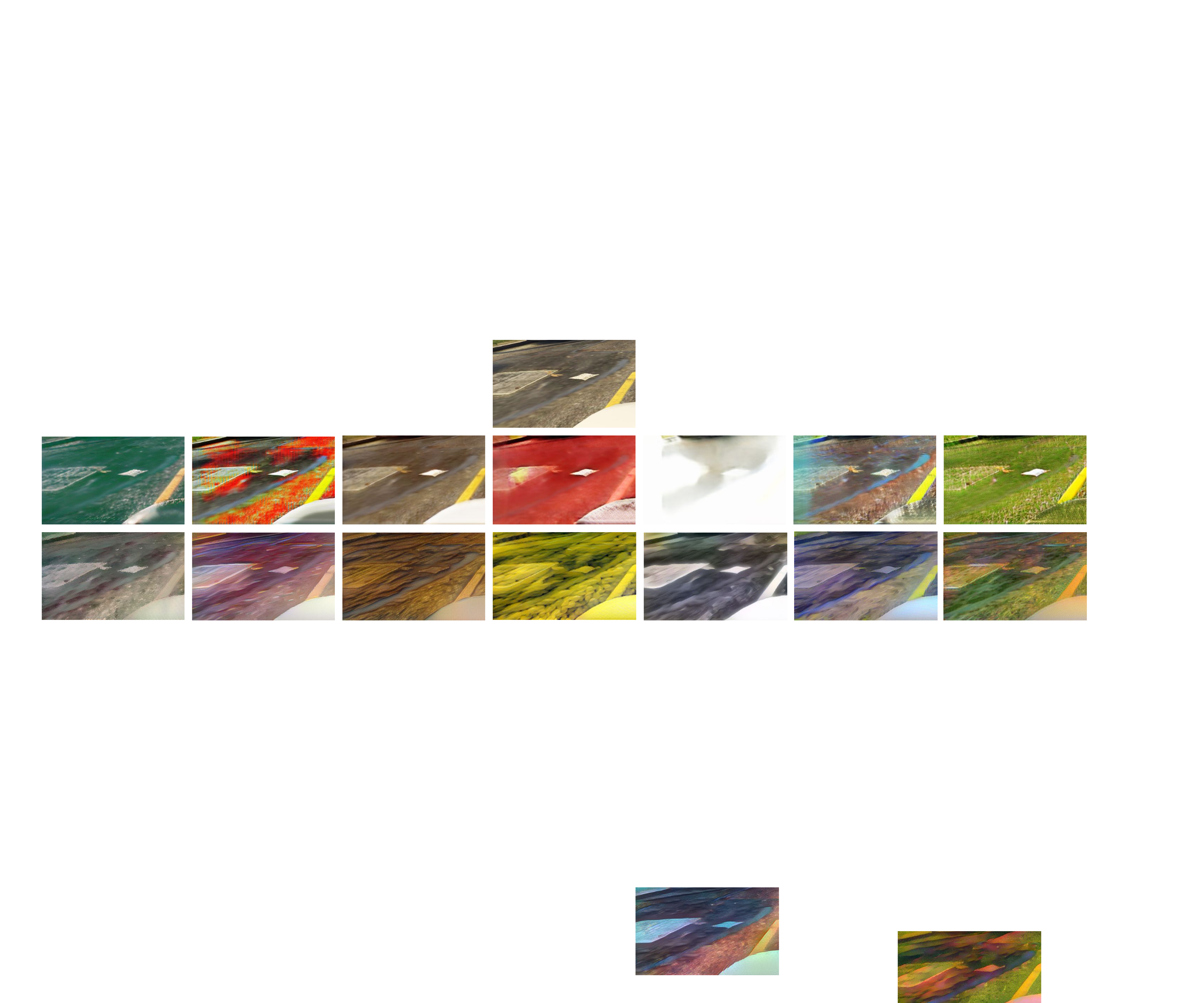}
    \caption{Texture comparison with DRPC. Cropped images from corresponding images from Figure \ref{fig:DRPC_images}.}
    \label{fig:road_texture}
\end{figure*}

Second, similar to our method, DRPC \cite{yue2019domain} uses CycleGAN to randomize source images. In Figure \ref{fig:DRPC_images}, we shows generated images using ImageNet classes used in \cite{yue2019domain} as auxiliary domains. Figure \ref{fig:road_texture} shows cropped images from Figure \ref{fig:DRPC_images}. 

In Figure \ref{fig:road_texture}, CycleGAN match auxiliary domain's color distribution while preserving the original image's synthetic texture. On the other hand, Style-swap diversifies texture. This is the most differentiated point of our method compared to DRPC. Since the main difference between the synthetic and real domains is not color but the texture, our texture-based method is more suitable than DRPC for randomization in synthetic to real tasks.

Also, our method is computationally more efficient than DRPC. Since training CycleGAN is a very costly process, DRPC only uses 15 auxiliary domains. On the other hand, since Style-swap does not require additional training for each style, it can handle many styles more easily. Hence our stylized datasets consist of 11,026 styles. 

Additionally, DRPC used Pyramid Consistency across Domain (PCD) loss to learn style-invariant feature. Because of this loss, a computation that is linearly proportional to the number of domains is required to simultaneously forward images across domains. Since DRPC used 16 domains, it requires at least 16 times more memory and computing power. 

Though DRPC used 16 domains, it might be required to consider more domains for more style-invariant representation, which demands impractical computation especially when the input’s resolution is large like GTA5 (1914x1052) and SYNTHIA (1280x760). On the other hand, our method requires a fixed amount of computation regardless of the number of styles.

\subsection{Ablation study}
We conduct an ablation study on Stage 1 in Table \ref{tab:effect of stylized dataset}. We divide the table into two sections according to the usage of adversarial loss.

In first section, \textit{Original source only} means training the segmentation network only with the original GTA5 images. \textit{Stylized source only} and \textit{Translated source only} use generated dataset  by Style-swap \cite{chen2016fast} and CycleGAN \cite{zhu2017unpaired} respectively. Results show model trained on only stylized source dataset outperforms other pixel-level \cite{wu2018dcan,gong2019dlow} method. 

The second section shows the results with the output-level adversarial training \cite{tsai2018learning}. Overall performances are improved compared to the first section.
Results show using both types (\textit{Stylized source} and \textit{Translated source}) by forwarding images alternately is better than only using \textit{Stylized source} images. This is because, while learning texture-invariant representation through the stylized images, the translated images guide the model to the target style. Following the results, we choose to use both \textit{Stylized source} \textit{Translated source} images with the output-level adversarial loss for the training segmentation network in Stage 1. 

We also conduct the ablation study for Stage 2 in Table \ref{tab:effect of self-training}. The results show in the third iteration of Stage 2 the segmentation model converged. Therefore we take three iterations for all results in Table \ref{tab:gtatocityscapes}, \ref{tab:synthia2cityscapes}.

\begin{table}[h!]
\begin{center}
\caption{Ablation study on Stage 1.}
\begin{tabular}{c c}
\hline
\multicolumn{2}{c}{GTA5 $\rightarrow$  Cityscapes}\\
\hline
method & mIoU \\
\hline
Original source only & 36.6 \\
DCAN \cite{wu2018dcan} & 38.5\\
Translated source only & 41.0 \\
DLOW \cite{gong2019dlow} & 42.3 \\
Stylized source only & \textbf{42.5} \\
\hline
Original source + Adv loss \cite{tsai2018learning} & 41.4 \\
Translated source + Adv loss \cite{li2019bidirectional} & 42.7  \\
Stylized source + Adv loss & 43.2 \\
Stylized/translated source + Adv loss & \textbf{44.6} \\
\hline
\end{tabular}
\label{tab:effect of stylized dataset}
\end{center}
\end{table}

\begin{table}[h!]
\begin{center}
\caption{Ablation study on Stage 2. In Stage 2-X, X means the number of iteration of self training.}
\begin{tabular}{c c}
\hline
\multicolumn{2}{c}{GTA5 $\rightarrow$  Cityscapes}\\
\hline
method & mIoU \\
\hline
Stage 1 & 44.6 \\
Stage 2-1 & 48.6 \\
Stage 2-2 & 50.2 \\
Stage 2-3 & 50.2 \\
\hline
\end{tabular}
\label{tab:effect of self-training}
\end{center}
\end{table}

\subsection{Robustness test}
To verify the texture-invariance of a model trained on the stylized dataset, we test the model on perturbated validation sets distorted by various noises. If the model is texture-invariant, it will be more robust to noises than other texture-dependent models. We generate noisy Cityscapes validation sets with noises that do not distort the shape of the original image's object. Following the method of \cite{hendrycks2019benchmarking}, we add Gaussian, Impulse, Shot and Speckle noise to the validation set. 

\begin{figure}[h]
\centering
\begin{subfigure}{0.23\textwidth}
\centering
\includegraphics[width=\textwidth]{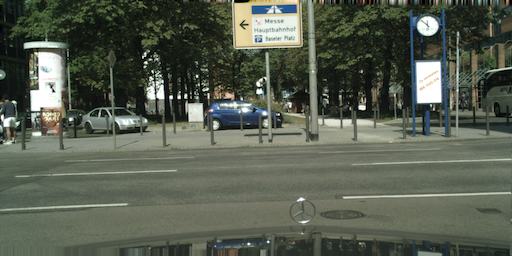}
\caption{Original image}
\end{subfigure}     
\begin{subfigure}{0.23\textwidth}
\centering
\includegraphics[width=\textwidth]{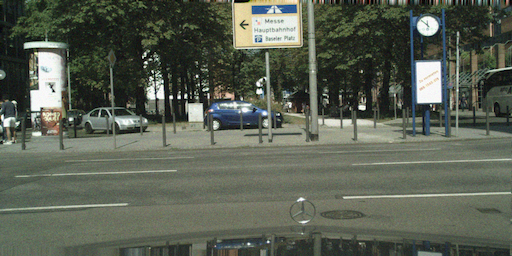}
\caption{Image with Gaussian noise}
\end{subfigure}     
\begin{subfigure}{0.15\textwidth}
\centering
\includegraphics[width=\textwidth]{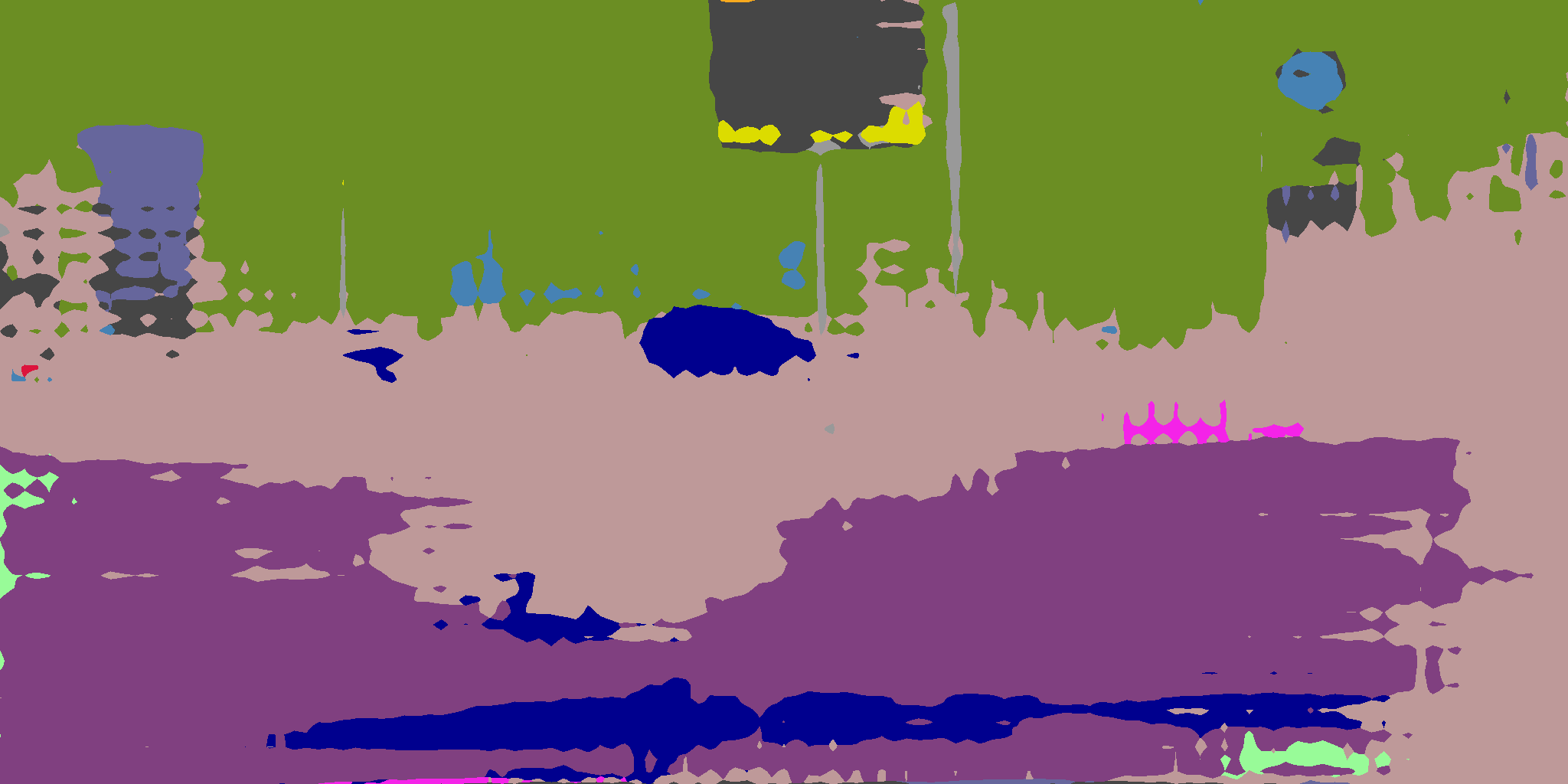}
\caption{AdaptSegNet}
\end{subfigure}     
\begin{subfigure}{0.15\textwidth}
\centering
\includegraphics[width=\textwidth]{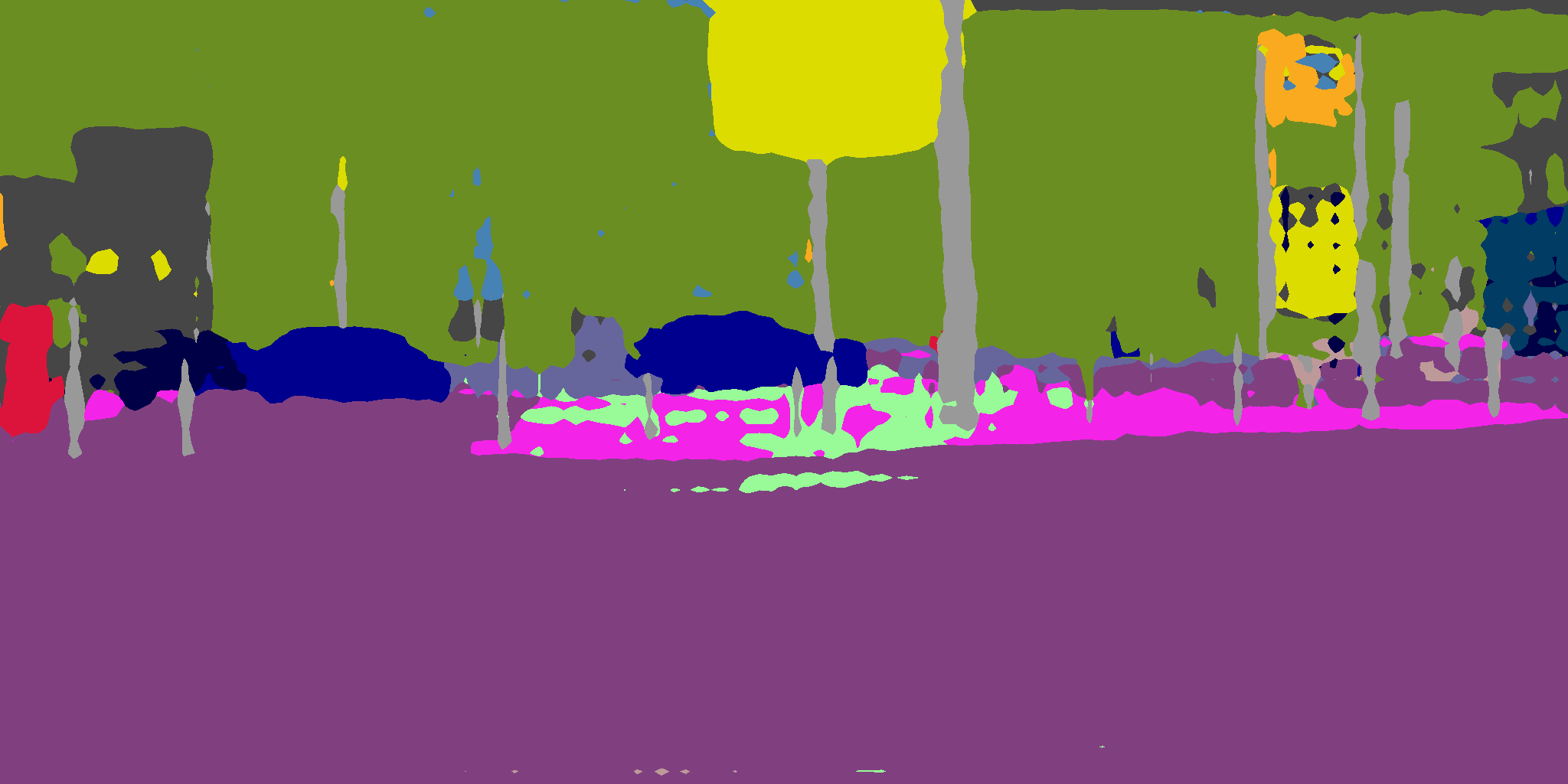}
\caption{Ours}
\end{subfigure}
\begin{subfigure}{0.15\textwidth}
\centering
\includegraphics[width=\textwidth]{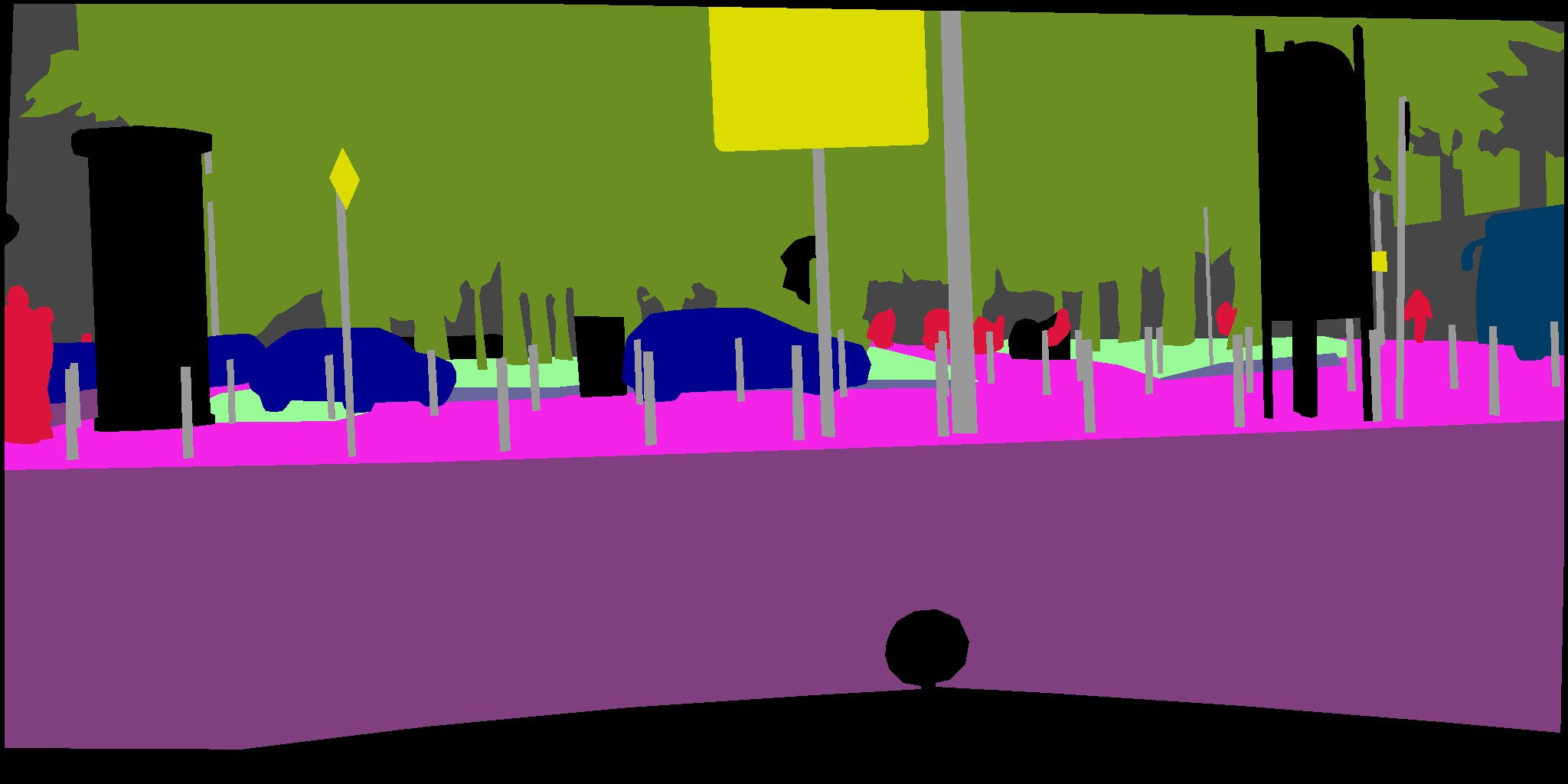}
\caption{Ground Truth}
\end{subfigure} 
\caption{Results on the validation image with Gaussian noise.}
\label{fig:noise_comparison}
\end{figure}

Results in Table \ref{noise_test} and Figure \ref{fig:noise_comparison} show that our model is much more robust to various noises than AdaptSegNet \cite{tsai2018learning} which is trained on original synthetic images. 

\begin{table}[h]
\begin{center}
\caption{Results on original and noisy validation set.}
\begin{tabular}{ccc}
\hline
    Method & AdaptSegNet\cite{tsai2018learning} & Stylized source only\\
    \hline
    Original & 42.4 & 42.5 \\
    Gaussian & 22.2 & \textbf{35.1} \\
    Impulse & 20.9 & \textbf{32.6} \\
    Shot & 24.9 & \textbf{38.2} \\
    Speckle & 32.5 & \textbf{41.1} \\
\hline
\end{tabular}
\label{noise_test}
\end{center}
\end{table}

\subsection{Qualitative results}
\begin{figure}[h]
    \centering
    \includegraphics[width=1\linewidth]{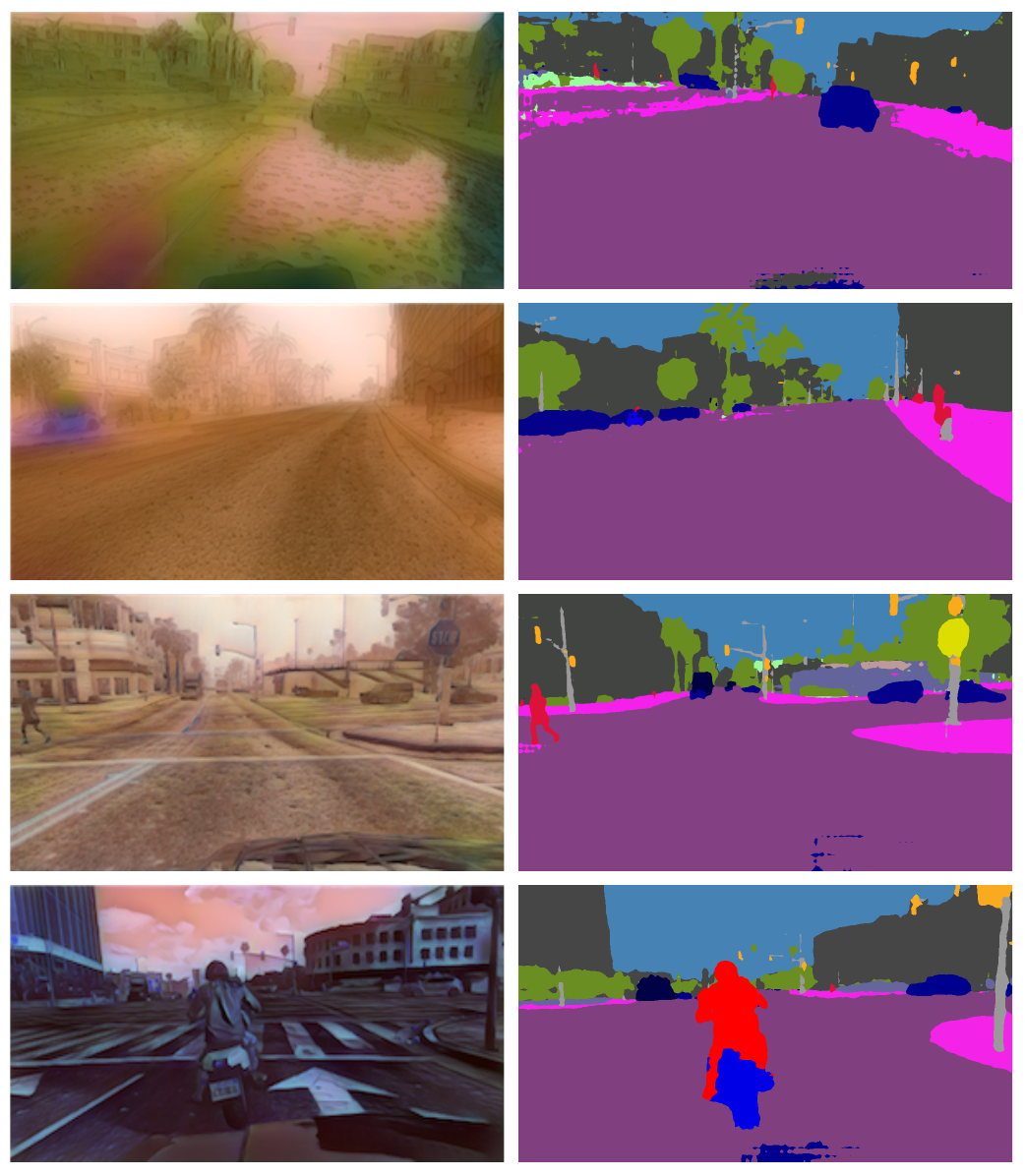} 
    \caption{Results on images with various texture. Images from the Stylized GTA5 (left column) and segmentation results (right column).}
    \label{fig:shape-dependency}
\end{figure}

To qualitatively demonstrate the texture-invariance of our model, we present segmentation results on images with various texture from the stylized source dataset in Figure \ref{fig:shape-dependency}. Results show our model is robust to texture variation.

\section{Conclusion}
In this paper, we present a method to adapt to the target texture. Using a style transfer algorithm, we generate the Stylized GTA5/SYNTHIA. The various texture of the stylized datasets works as a regularizer to make the segmentation model learn texture-invariant representation. We show the texture-invariance of our model qualitatively on images with various texture and quantitatively on noisy validation sets. Based on the texture-invariant representation, we use self-training to get direct supervision of the target texture. Experimental results show the effectiveness of our approach, which achieves new state-of-the-art performance in the GTA5 to Cityscapes benchmark. Besides, we analyze the influence of texture across different classes. Also, we compare our style transfer-based method and CycleGAN-based methods in terms of reducing the texture gap between the synthetic and real domains.

\section{Acknowledgement}
This work was supported by the National Research Foundation of Korea grant funded by Korean government (No. NRF-2019R1A2C2003760).

{\small
\bibliographystyle{ieee_fullname}
\bibliography{egbib}
}

\end{document}